\pdfoutput=1
\documentclass[11pt,a4paper]{article}
\usepackage[T1]{fontenc}
\usepackage[utf8]{inputenc}
\usepackage[]{acl}
\usepackage{microtype}
\usepackage{danuedfs}
\usepackage{times}
\usepackage{latexsym}
\usepackage{graphicx}
\usepackage{amsmath, amsfonts, amssymb, amsthm, mathtools}
\usepackage{dsfont}
\usepackage{booktabs}
\usepackage{algorithm, algorithmic}

\usepackage{booktabs}
\usepackage{microtype}

\usepackage{xspace}
\usepackage{todonotes}

\makeatletter
\if@todonotes@disabled

\else

\fi
\makeatother

\title{Position-based Prompting for Health Outcome Generation}
\author{Micheal Abaho$^1$
  \\\And
  Danushka Bollegala$^{1,2}$\Thanks{ Danushka Bollegala holds concurrent appointments as a Professor at University of Liverpool and as an Amazon Scholar. This paper describes work performed at the University of Liverpool and is not associated with Amazon.}\\
    \kern+10em $^1$University of Liverpool, United Kingdom \\
     \kern+10em $^2$Amazon \\
  \kern+10em \texttt{\{m.abaho,danushka,prw,shinds\}@liverpool.ac.uk} 
  \\\And
  Paula R Williamson$^1$ \\
  \\\And
  Susanna Dodd$^1$ \\
  \\}
\date{}

\begin{document}
\maketitle

\begin{abstract}
Probing factual knowledge in Pre-trained Language Models (PLMs) using prompts has indirectly implied that language models (LMs) can be treated as knowledge bases. To this end, this phenomena has been effective, especially when these LMs are fine-tuned towards not just data, but also to the style or linguistic pattern of the prompts themselves. We observe that, satisfying a particular linguistic pattern in prompts is an unsustainable, time-consuming constraint in the probing task, especially because, they are often manually designed and the range of possible prompt template patterns can vary depending on the prompting task. To alleviate this constraint, we propose using a position-attention mechanism to capture positional information of each word in a prompt relative to the mask to be filled, hence avoiding the need to re-construct prompts when the prompts' linguistic pattern changes. Using our approach, we demonstrate the ability of eliciting answers (in a case study on health outcome generation) to not only common prompt templates like Cloze and Prefix, but also rare ones too, such as Postfix and Mixed patterns whose masks are respectively at the start and in multiple random places of the prompt. More so, using various biomedical PLMs, our approach consistently outperforms a baseline in which the default PLMs representation is used to predict masked tokens.
\end{abstract}

\section{Introduction}
Language models (LMs) as knowledge bases (KBs) (LM-as-KB) is a rapidly growing phenomenon attracting a lot of attention in the Natural Language Processing (NLP) community \cite{petroni-etal-2019-language, brown2020language, shin2020autoprompt, schick2020s}. LM-as-KB implies the usage of LMs as an alternative or at least a proxy for explicit KBs. To achieve LM-as-KB, researchers adopt prompt-based learning (PBL) in which LMs learn to probabilistically predict missing information once given fill-in-the-blank prompt inputs \cite{liu2021pre} such as ``Eiffel tower is located in \underline{\hspace{0.5cm}}''. PBL has generally been a success, for example, in a systematic survey of prompting methods, \newcite{liu2021pre} indicate that \textit{``pre-train, prompt and predict''} is a new paradigm replacing \textit{``pre-train and fine-tune''} paradigm in NLP. Because of this success, the rationale that LMs contain factual retrievable knowledge (LM-as-KB) is ostensibly justified and therefore continually explored.  

\begin{figure}[t]
    \centering
    \includegraphics[width=1.02\columnwidth]{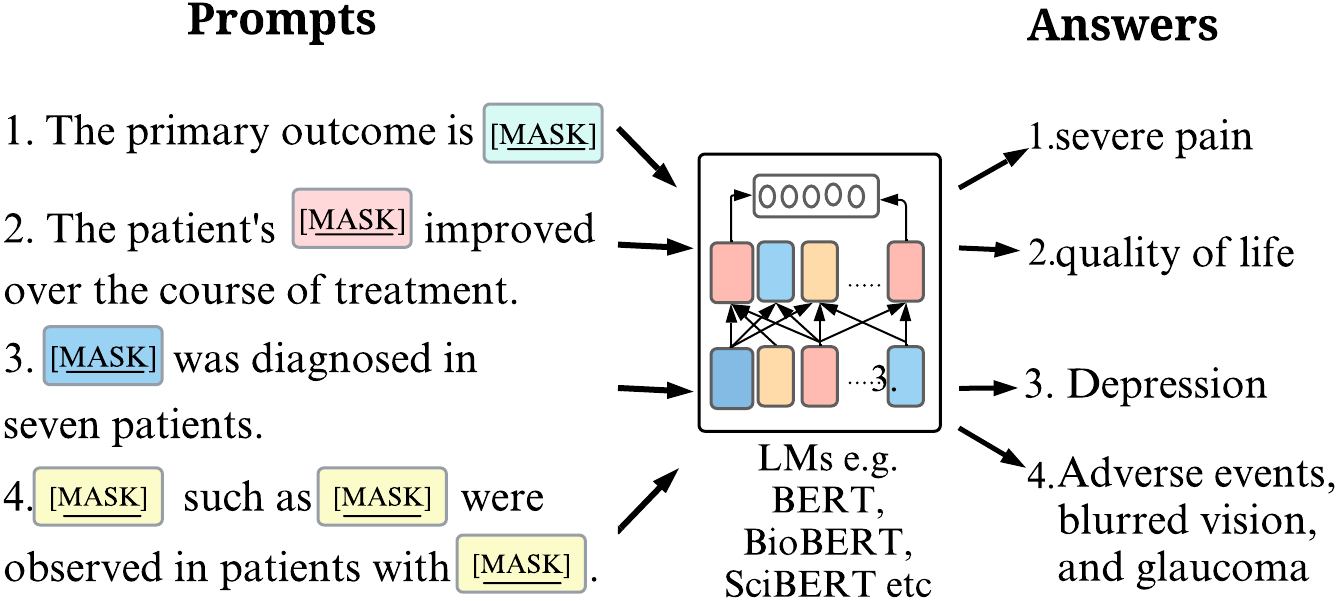}
    \caption{Prompt query variants used for probing evidence (in form of health outcomes) from PLMs, including common styles like Prefix (1) and Cloze (2) style, as well as rare styles Postfix (3) and Mixed (4) styles with [MASK] token/s at the beginning and in multiple positions in the prompt.}
    \label{fig:prompt_patterns}
    \vspace{-1em}
\end{figure}

The prompt sequences often used in PBL have a masked token or span (denoted by [MASK] in the remainder of the paper) that positionally appears either in the middle (Cloze-style) \cite{petroni-etal-2019-language, schick2020s, cui2021template} or at the very end of the sequence (Prefix style) \cite{qin2021learning, shin2020autoprompt}. Moreover, we learn that the majority of the PBL tasks probe relational knowledge possessed by pre-trained language models (PLMs) \cite{jiang2020can, petroni-etal-2019-language, davison2019commonsense}, which implies that the prompt inputs used in querying the PLMs have to contain relational information (such as \textit{``subject-relation-object''} triples). Furthermore, we observe that, a fair amount of time in several PBL tasks is spent reconstructing prompt inputs through manually designing templates \cite{petroni-etal-2019-language, davison2019commonsense} or corrupting prompt inputs through deletion \cite{lewis2019bart}, replacement \cite{raffel2019exploring} or permutation \cite{heinzerling2020language}.

As discussed above, we notice that, the syntactic and semantic structure of prompt inputs is a constraint encountered in PBL, notwithstanding the multitude of 
constraints that could arise given that PBL is inherently a text generation task \cite{liu2021pre}. This constraint will usually require researchers to laboriously prepare supervised data with prompts whose linguistic patterns suit the objective of the prompting task, For instance, \cite{davison2019commonsense, jiang2020x, heinzerling2020language}, use templates that reformulate prompts to contain relational information connecting a particular text span to the to-be filled information. However, template-based prompt reformulation has two main challenges. First, it presents a risk of corrupting the grammar of the prompts unwittingly \cite{davison2019commonsense}.
Second, the search space of the candidate prompts is too large \cite{gao2020making} and is practically impossible to create templates that can enumerate all possible linguistic patterns that prompt queries can be tailored to.
For example, prompt template patterns with missing information at the beginning and or with multiple missing information in a sequence are yet to be explored in prior works. 

To address the above-mentioned challenges, we propose a strategy we denote position-based prompting (PBP), which is less concerned about the linguistic pattern or shape the prompt takes on, but rather focuses on the words (that the prompts are composed of) and their positions relative to the [MASK]. PBP is focused on shifting the emphasis on subject-relation-object triples to the masked positions as well as the interaction of all the other words with the [MASK]s position. PBP is built to automatically adjust from one prompt template to another, which essentially eliminates the need to prepare hand crafted prompts in the event that an LM is to be probed for rare knowledge. In its architecture, PBP enhances contextualised word representations with position-aware representations to solve fill-in-the-blank tasks. In our approach, we fine-tune PLM parameters along with position-oriented parameters to generate position-based contextualised word representations. 

To test our approach, we investigate how well biomedical LMs store and recall information relevant to biomedical entities, with a specific interest in health outcomes, which are defined as measurements or observations used to capture and assess the effect of treatments \cite{Williamson2017The1.0}.    
In addition to the Prefix and Cloze styles, we incorporate two rare prompt style patterns that we denote Postfix and Mixed, where the former contains the [MASK] token/s at the beginning of the prompt sequence and the latter has multiple [MASK] token/s in various positions (\autoref{fig:prompt_patterns}). Our approach obtains mean scores (across several biomedical LMs) in Exact Match (EM) and Partial Match (PM) metrics that are an improvement (2.4\% across both metrics) over those obtained using the vanilla PLM representations, reporting a significant improvement of 6.49\% in F1 on the EBM-NLP \cite{Nye2018ALiterature} dataset. As later defined in \autoref{sec:metrics}, EM measures the percentage of predictions of all [MASK] tokens (or spans) that match the ground truth, whereas PM measures the percentage of correctly predicted [MASK] tokens.

\section{Entity memorisation and recalling}
Large-scale LMs with billions of parameters have already shown to recall facts that were observed in the training data \cite{heinzerling2020language, jiang2020x}. 
However, the ground truth for these  LMs to achieve this is already laid with systematically handcrafting rules to follow in creating the prompt input sequences they receive at the training stage. 
For instance, the majority of the prompts created in PBL tasks embed knowledge in form of triples \{\textit{subject,relation,object}\} such that LMs could correctly predict \textit{object} entities when prompted with a sequence containing a \textit{subject} and \textit{relation} or otherwise predict \textit{subject} entities when prompted with a sequence containing an \textit{object} and a \textit{relation} \cite{sung2021can, jiang2020x, qin2021learning}. Whichever the case, models often predict answers as shown in \eqref{eqn:argmax}.
\begin{align}
    \label{eqn:argmax}
    \hat{y}_i = \underset{y_i}{\argmax} \ \ p([{\rm MASK}]=y_i|x_{prompt}) 
\end{align}
where $i$ is the position of masked token within a prompt $x_{prompt}$.

In this work, we however do not assume any prior knowledge contained in a prompt, but rather simply locate outcome entities in the sentences extracted from Randomised Clinical Trial (RCT) abstracts and mask them, an approach we refer to as \emph{custom masking}. 

\section{Method}
\vspace{-0.5em}
In addition to formally defining the task we undertake, this section discusses the data used as well as the different stages of our proposed PBP strategy.

\vspace{-0.2em}
\subsection{Task}\label{sec:Task}

Let us consider an input prompt sequence $s$ with one or more outcomes masked such that $s=x_1, \ldots [M]_i \ldots [M]_j \ldots x_n$, where $[M]$ is a masked token sequence, $[M] = \{x_i\}^{i+|M|}_{i\geq1}$ , $i \in [1,n]$ and $|M|$ is the length of the masked sequence. We consider four different prompt query variants shown in \autoref{fig:prompt_patterns}:
\textbf{Prefix prompts} contain $[M]$ at the end of the prompt, \textbf{Cloze prompts} contains $[M]$ in the middle of the prompt, \textbf{Postfix prompts} contain $[M]$ at the start of the prompt, and \textbf{Mixed prompts} where there are several masked sequences distributed across the prompt. 
The questions we then pose are: (a) \emph{can we determine how knowledgeable biomedical PLMs are of stored facts such as health outcomes?}, and 
(b) \emph{If queried with any of the above variants, would these PLMs correctly fill in $[M]$s with the correct outcomes?}

\subsection{Datasets}
\label{sec:datasets}
Different from previous PBL works, we neither create custom templates nor do we reformulate prompts to follow an ideal linguistic pattern. We use plain raw sentences (that mention health outcomes) extracted from RCT PubMed abstracts, which are contained in the revised version of EBM-NLP \cite{abaho2019correcting} and EBM-COMET \cite{abaho2021assessment} datasets. Both of these datasets support evidence based medicine (EBM) tasks such as extraction of health outcomes from clinical trials \cite{beltagy2019scibert, abaho2021detect}.

We do not eliminate any of the abstract sentences that do not mention outcomes, because we aim to familiarise the PLM (at fine-tuning) with text or context in RCT abstracts which generally report about outcomes during clinical trial studies \cite{Williamson2017The1.0}. We refer to these sentences as \emph{no\_blank sequences} and use them alongside the prompt query variants introduced earlier. 
To our advantage, several sentence segments have no outcome annotations in both the EBM-NLP and EBM-COMET datasets.

\subsection{Masked Language model and Prompt engineering}
We extract a hidden state $h_i$ for each token in an input prompt $s$ using a domain-specific PLM,
\begin{align}
\vec{h_i} = \mathrm{PLM_\theta}(x_i)
\label{eqn:plm}
\end{align}
where $\vec{h_i}$ is a hidden state for the word $x$ at position $i$. The matrix of hidden states for the entire input prompt is represented as $\mat{H} \in \R^{n \times k}$, where $n$ is number of words in $s$ and $k$ is the hidden state size.

We define a function $f_{\textbf{prompt}}$ that concatenates the $h_i$ in \eqref{eqn:plm} to a randomly initialised $d$ dimensional vector, which we denote as $z_t$ corresponding to one of the four prompt query variants or the additional \emph{no\_blank sequences} (introduced in \textsection\ref{sec:datasets}), where $t \in \mathrm{[\textit{prefix, cloze, postfix, mixed, no\_blank}]}$. The function ensures that if an input $s$ is a Prefix prompt, the corresponding vector $z_{prefix}$ is concatenated to each $h_i$ generated from $s$ as shown in \eqref{eqn:concat}. This is done to enable knowledge transfer from one prompt query to another. For example, Mixed prompts are by construction a combination of Prefix, Postfix, and Cloze, hence they should benefit from information sharing via a common vector space.

\begin{align}
 f_{\textbf{prompt}}(h_i) = [z_t;h_i]
 \label{eqn:concat}
\end{align}
$z_t \in \R^{d_t}$, where $z_t$ is a query type embedding of size $d_t$. 

\subsection{Position based conditioning (PBC)}
\label{section:pbc}
To enrich the token representations, we propose a position-based attention mechanism to steer the model's focus on relevant information in the input prompt. We define a sequence of position ids for each input prompt, where all masked positions take on an id of 0 and all the other tokens take id's relative to the masked position id. For example given a Cloze prompt with $m$ tokens, we assign a mask at position $i$ an id 0, and resulting sequence of position ids is $p = [1-i, 2-i, \ldots, -1, 0, 1, \ldots, (m-1)-i, m-i]$. 
We compute an attention vector $A^{(s)}$, given by \eqref{eq:A}, for an input prompt $s$ that allows each token to interact with every other token and retain knowledge of the relative position of the masked tokens in the input sequence. 
\begin{align}
\label{eq:A}
    \vec{A^{(s)}} =  \mathrm{softmax}(\mat{V}^{\T}\tanh(\mat{W}\mat{H}^{\T} + \mat{U}\mat{P}^{\T}_s))
\end{align}
Here, $\vec{A^{(s)}} \in \R^{n \times 1}$, $\mat{V} \in \R^{k_a \times 1}$, $k_a$ is size of attention layer, $\mat{W} \in \R^{k_a \times k}$, $\mat{P}_s \in \R^{n \times k_p}$ and $\mat{U} \in \R^{k_a \times k_p}$. $\mat{P}_s$ is a matrix of position embeddings of size $k_p$ extracted for each position $p_n$ in the input prompt $s$. These embeddings are extracted from a trainable matrix $\mat{P} \in \R^{2n \times k_p}$ of randomly initialised vectors of size $k_p$ for all possible positions $2n$ where $n$ is the maximum sequence length, $|\{p_n\}_{-n}^{n-1}| = 2n$. The position based representation of each token is then computed with respect to the type of prompt. For the Prefix, Postfix and Cloze prompts, we obtain a prompt representation $M^{s}$ given by \eqref{eq:M}. 
\begin{align}
\label{eq:M}
    \mat{M}^{(s)} =  \vec{A}^{(s)}\mat{H}
\end{align}
Here, $\mat{M}^{(s)} \in \R^{n \times k}$. 
For the Mixed prompts in which we have multiple masked positions within the input sequence, 
we avoid biasing the attention mechanism towards masks at a specific position and thereby considering as many position id sequences as there are masked positions in the input prompt. For example, given a sequence with 3 masked positions, $s = [M],x_2,x_3,[M],x_5,x_6,[M]$, we obtain 3 position id sequences, i.e. the combined position id sequences is,
\begin{align*}
\label{eq:state2}
   P^{(s)} = \underset{i}{\bigcup}P_i ,
\end{align*}
where each $P_i$ is obtained with respect to the current mask position $i$.
For the example above, we have $P^{(s)}$ = \{[0,1,2,3,4,5,6], [-3,-2,-1,0,1,2,3,], [-6,-5,-4,-3,-2,-1,0]\}, where the first position id sequence is obtained by treating the $[M]$ at position 1, as mask at $i$, the second is obtained by treating the $[M]$ at position 4 as mask at $i$ and finally the third by treating $[M]$ at the last position as mask at $i$. 
Attention vectors are computed for each position id sequence $(P_i)$ and subsequently used to obtain the prompt representation $\mat{M}^{s}_{P_i}$. We compute the final representation of a Mixed prompt as the mean pool across these different representations,
\begin{align}
\mat{M}^{(s)} = \sum_i^{|P^{(s)}|}\mat{M}^{s}_{P_i}
\end{align}

\begin{table*}[t]
\centering
\resizebox{2.05\columnwidth}{!}{
\begin{tabular}{@{}lcccccccccccc@{}}
\toprule Dataset- & \multicolumn{6}{c}{EBM-COMET} & \multicolumn{6}{c}{EBM-NLP}\\  \midrule
         Method-       & \multicolumn{2}{c}{Baseline} & \multicolumn{2}{c}{PBC} & \multicolumn{2}{c}{Contextual PBC} & \multicolumn{2}{c}{Baseline} & \multicolumn{2}{c}{PBC} & \multicolumn{2}{c}{Contextual PBC}\\ \midrule
         Metric-       & EM                                        & PM                                       & EM                                             & PM                                             & EM           & PM     & 
         EM                                        & PM                                       & EM                                             & PM                                             & EM           & PM
         \\ \midrule \midrule
BERT            & 43.12                                     & 47.55                                    & 43.04                                          & 49.84                                          & 44.32        & 55.94   & 37.40 & 45.55 & 41.10 & 47.00 & 47.31   & 51.06   \\
BioBERT         & 50.71                                     & 58.01                                    & 50.55                                          & 58.61                                          & 53.34        & 59.65  & 51.15 & 55.62 & 51.19 & 53.80 & 52.15 & 54.50      \\
SciBERT         & 61.17                                     & 67.48                                    & 62.34                                          & 69.85                                          & 63.00        & 70.95   & 57.12 & 62.25 & 57.18 & 63.75 & 59.44 & 63.91       \\
Biomed\_RoBERT{\small A} & 44.01                                     & 59.67                                    & 44.32                                          & 59.73                                          & 44.32        & 62.86 & 40.45 & 51.72 & 47.21 & 49.81 & 49.17 & 55.00     \\
UmlsBERT        & 31.05                                     & 34.61                                    & 30.47                                          & 35.77                                          & 31.88        & 36.46    & 28.66 & 33.15 & 30.02 & 38.51 & 39.16 & 40.15     \\
\midrule
Mean score        & 46.01                                     & 53.46                                    & 46.14                                         & 54.76                                          & 47.37       & 57.17    & 42.96 & 49.66 & 45.34 & 50.57 & 49.45 & 52.92    \\\bottomrule
\end{tabular}
}
\caption{Table reports EM and PM accuracies of the various biomedical Pre-trained Language Models for the outcome recalling experiments. Mean score in a particular column is the average across all results in that column.}
\label{tab:main-results}
\vspace{-1em}
\end{table*}

\subsection{Prompt fine-tuning}
The predicted probability of each vocabulary token is estimated via \eqref{eq:prob}.
\begin{align}
\label{eq:prob}
    y =  \mathrm{softmax}(f(W_v\mat{M}^{(s)^{\T}})
\end{align}
Therein, $W_v \in \R^{v^* \times k}$, $v^*$ is the vocabulary size and $f$ is a non-linear activation function.
We use a BERT-based loss in predicting the masked tokens in each input given by \eqref{eq:loss}.
\begin{align}
\label{eq:loss}
 \vec{L}_{PLM} = -\sum_{s \in \cT}\sum_{i}^{n} \log P(y_i|s)
\end{align}
where $\cT$ is the set of training example prompts.
Some of the prompt query variants (Postfix and Prefix) are rare in the datasets, and some other prompt sequences are quite lengthy. This poses a challenge particularly when using small PLMs (with few parameters) to recall factual information. In order to mitigate model forgetfulness in such examples, we introduce an auxiliary task that computes a text classification loss as a cross entropy loss given by \eqref{eq:CE}.
\begin{align}
\label{eq:CE}
 \vec{L}_{TC} = -\sum_{s \in \cT}\sum_{i \in n} \log P(y_i|y_{<i},s)
\end{align}
 The overall training loss is defined as the weighted combination of the two losses as given in \eqref{eq:total-loss}.
\begin{align}
\label{eq:total-loss}
 \vec{L} = \vec{L}_{PLM} + \lambda \vec{L}_{TC}
\end{align}
Similar to \cite{chronopoulou2019embarrassingly} and \cite{schick2020exploiting}, we introduce a weighting parameter $\lambda (>0)$ to adapt the auxiliary losses to the main mask prediction task.  

\subsection{Prediction}
\label{sec:prediction}
Similar to BERT \cite{devlin2018bert}, we consider generating outputs in parallel, initially treating the default representations provided by the model in \eqref{eqn:plm} as a baseline and therefore use them to predict tokens in masked positions. 
We then use position-aware representation obtained using the attention mechanism in \textsection \ref{section:pbc} to predict the mask tokens, calling these results Position-based conditioning (PBC). Lastly, we endeavour to retain the contextual knowledge presented by the PLMs as much as we possibly can by computing an average of the Baseline and PBC representations and term these Contextual PBC.

\section{Experiments}
 In our experiments, we use several PLMs that are pre-trained on clinical texts such as PubMed abstracts, which often report outcomes such as BioBERT \cite{lee2020biobert}, SciBERT \cite{beltagy2019scibert} and Biomed\_RoBERT{\small A} \cite{gururangan-etal-2020-dont}. Additionally, we include UmlsBERT because it augments BERT's pre-training input with semantic type embeddings aligned to clinical knowledge (semantic types) in the Unified Medical Language System (UMLS) Metathesaurus \cite{michalopoulos2020umlsbert}.
 We also use BERT~\cite{devlin2018bert} as a vanilla PLM that has not been pre-trained specifically on clinical texts.
 
 \subsection{Training and Evaluation}
 Unlike previous works where a particular relation within a prompt e.g. \textsf{born-in}, \textsf{lives-in} etc. might appear multiple times within the train set, in our case, prompts are not semantically related in any way (i.e. their is no relation knowledge that can be transferred over from one prompt to another). 
 Because of the nature of our prompts, we believe it might be harder for the model to memorise them, we therefore opt to train the models until the perplexity on the training data reaches 1 or until the accuracy on the validation data saturates. 
We examine the model's generalisation ability to transfer knowledge to unseen prompts in few-shot and zero-shot settings. 
 For the few-shot setting, we design experiments where we measure a model's accuracy in generating outcomes (as answers), which it encountered in a small number of prompts during training. 
 The contexts in these evaluation prompts are not encountered during training.
 For example, consider an evaluation prompt -- \textit{``The patient's overall [MASK] improved according to the HRQOL questionnaire}'', the model would not have encountered the context surrounding the \textit{``[MASK]''}. 
 For the zero-shot evaluation, the model would have neither encountered the prompt nor the target outcomes during training. 
 To simulate both the zero- and few-shot settings, we randomly split the datasets into train (80\%) and test (20\%) splits, and use the latter for the generalisation evaluation task shown in \autoref{tab:fewshot}. 
 We tune all hyperparameters using the validation data, and obtain optimal values as follows: learning rate - 5e-5, batch size - 8, query type embedding size - 50, position embedding size - 300 and an attention layer size - 200. Further details on tuning bounds are provided in the Appendix.
 
\paragraph{Metrics:}
\label{sec:metrics}
We define two different metrics for evaluating the proposed PBP strategy: Exact Match (EM) and Partial Match (PM).
EM counts a prediction as 1 only if it matches completely with the correct answer, whereas PM uses the fraction of the overlapping tokens between the predicted and correct answers. Both EM and PM are averaged over all test instances to compute aggregated evaluation metrics, and we report their percentages in the paper.





\section{Results}
\label{sec:results}
In this section, we evaluate how well the model generates health outcomes when queried to answer a given prompt.
For example, ``\textit{After patients were given sorafenib, they reported [MASK]}'', the model should correctly generate the outcome  \textit{Fatigue} for the \textit{[MASK]}.

\subsection{Outcome memorisation and retrieval}
\autoref{tab:main-results} shows the performance of the proposed PBC method in the outcome generation task. 
As observed, PBC consistently outperforms the baseline across most of the clinically informed BERT LMs (for both datasets), particularly for the PM results. 
More interestingly, we notice that Contextual PBC further improves the performance (both in EM and PM), indicating the importance of preserving the contexts in the position-based representations. 

Comparing the different LMs, we found that, SciBERT performs best followed by Biomed\_RoBERT{\small A} and BioBERT. Since all tested models follow the original BERT's architecture, we hypothesize that, the nature of corpora used in pre-training the best performing models was responsible for the performance, i.e. unlike UMLSbert and BERT, all the other models are pre-trained on text that includes PubMed abstracts, which often report outcomes. 
Additionally, we observe that PM results were generally better than EM results, which we attribute to the fact that PM is less strict compared to EM because it rewards the model for correctly generating a few of the tokens in the masked positions.
Overall, the results suggest that PBC can be used to effectively retrieve facts such as health outcomes (biomedical entities) by simply augmenting contextual word representations with position-aware representations.    

\subsubsection{Prompt query variants}
\begin{table}[t]
\centering
\resizebox{0.95\columnwidth}{!}{
\begin{tabular}{@{}lcccc@{}}
\toprule
        & \#   & \begin{tabular}[c]{@{}c@{}}Average\\ prompt length\end{tabular} & EM    & PM    \\ \midrule \midrule
Postfix & 65   & 18.5 & 48.43 & 58.51 \\
Prefix  & 53   & 9.1  & 69.23 & 77.24 \\
Cloze   & 630  & 24.2 & 50.08 & 60.49 \\
Mixed   & 2594 & 38.8 & 43.68 & 45.46 \\ \bottomrule
\end{tabular}
}
\caption{Exact Match (EM) and Partial Match (PM) accuracies for Outcome memorisation/recalling for the different prompt types using the EBM-COMET dataset.}
\label{tab:prompt_query_variants}
\end{table}

In \autoref{tab:prompt_query_variants}, we notice that the accuracy with which a model correctly answers Prefix prompts is significantly higher than that of the other prompts. 
We attribute this performance to the short length of these spans such as the one shown in \autoref{tab:example_prompts} and the average number of tokens to decode per prompt. 
We also notice that the model struggles to correctly answer Mixed prompts compared to other types of prompts. 
We attribute this to the fact that, Mixed prompts are generally very long sequences (38.8 tokens on average) and contain multiple masked positions to be predicted. 

\subsection{Few- and Zero-shot Evaluations}

\begin{table}[t]
\centering
\resizebox{0.7\columnwidth}{!}{
\begin{tabular}{@{}lcccc@{}}
\toprule
       & Cloze & Mix & Postfix & Prefix   \\ \midrule \midrule
        \# & 174 & 613 & 13 & 12 \\
 \bottomrule
\end{tabular}
}
\caption{Number of prompts per prompt type used in evaluation of the few- and zero-shot settings.}
\label{tab:fewshot}
\vspace{-1em}
\end{table}

\begin{figure}[t]
    \centering
    \includegraphics[width=\columnwidth]{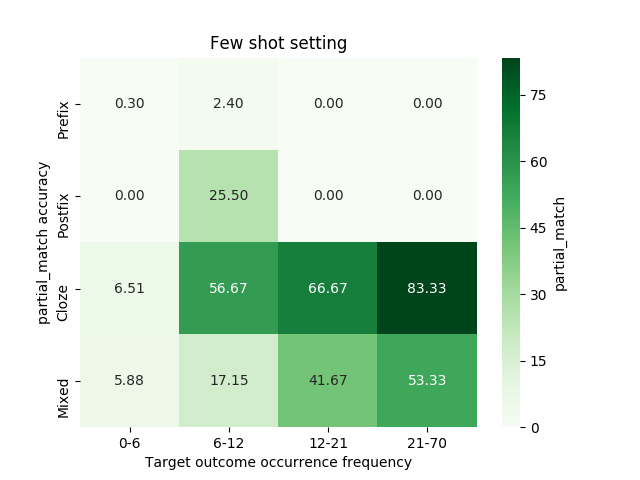}
    \includegraphics[width=\columnwidth]{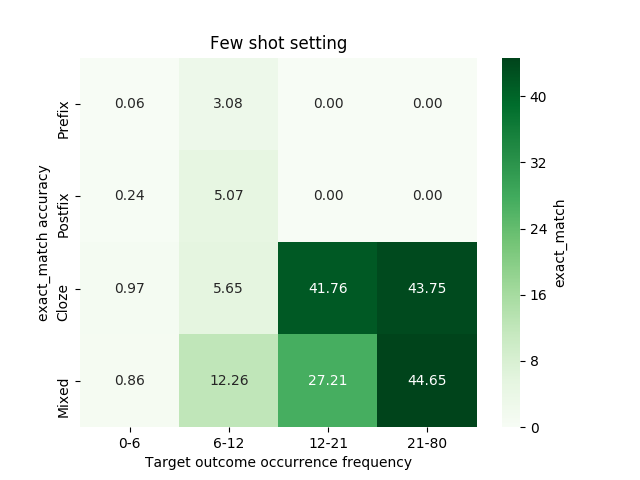}
    \caption{Visualizing the Partial Match and Exact match accuracies when the best model (SciBERT+Contextual PBC+EBM-COMET) is trained with only a certain number of target outcomes.}
    \label{fig:shot}
    \vspace{-1em}
\end{figure}

\begin{figure*}[t]
    \centering
    \includegraphics[width=\textwidth]{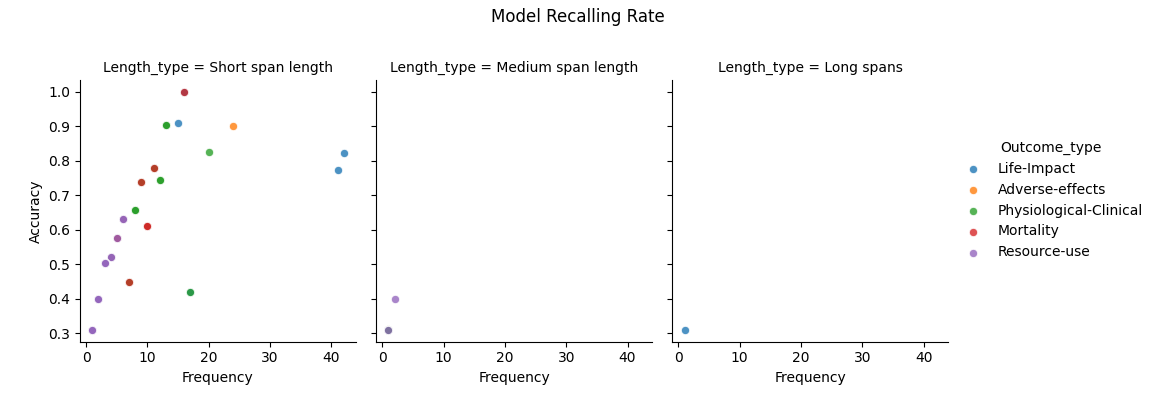}
    \caption{Analysis of the accuracy (PM) with which best model (SciBERT+Contextual PBC+EBM-COMET) recalls different types of factual information (outcome types) with varying span lengths and occurrence frequency (in the dataset).}
    \label{fig:freq_len}
    \vspace{-1em}
\end{figure*}

To evaluate the model's generalisability, we fine-tune the model towards a small amount of target outcomes, and then measure the transferability of this knowledge by requiring the model to accurately generate these outcomes in prompts with completely different contexts. 
Test set prompts in \autoref{tab:fewshot} are carefully chosen using regular expression matching such that the contexts surrounding the missing outcomes are different from that of similar outcomes observed during training. 
For example, the model could have been trained on the outcome ``adverse events'' in five different prompts, and then at evaluation, the model is required to generate the same outcome, however using prompts that are different from those encountered during training. By \emph{different} here we mean that the context (e.g. \{ctxt\} surrounding masks [M] in \autoref{tab:example_prompts}) in the prompt changes during this evaluation. 
\autoref{fig:shot} plots shows results of model evaluation on prompts (\autoref{tab:fewshot}). 
As observed in the plots, the model struggles to generate outcomes it hardly encountered during training (i.e. outcomes appearing in 0-6 prompts or 6-12 prompts). 
This is mostly evident in generating outcomes for Prefix and Postfix prompts, which is because there were not just few evaluated prompts of this types, but there were also few (53 and 65 respectively as shown in \autoref{tab:prompt_query_variants}) in the train set. 
However, we see a trend of performance improvement when the frequency of target outcomes encountered during training increases, particularly for the Mixed and Cloze prompt. 

\section{Analysis}
\subsection{Impact of Length and Frequency of Outcomes}
We partition the entire set of outcomes in EBM-COMET into 3 different groups based on lengths. Dividing the length of the longest outcome (22) by 3, we get approximately 7 which we use to create 3 groups i.e. 1) ``short span length'' to represent outcomes that are $\leq 7$ tokens long, 2) ``medium span length'' to represent outcomes of $7 >$ and $\leq 14$ tokens, and finally 3) ``long spans'' to represent outcomes of $\ge 14$ tokens long. \autoref{fig:freq_len} shows how well the best model (SciBERT+Contextual PBC+EBM-COMET) performs when recalling outcomes of varying lengths and frequencies. 
Following prior work on EBM NLP, we endeavour to show the model's outcome recall rate by outcome type, which can be informative in terms of the complexity of modelling these outcomes. We firstly notice the skewed distribution of outcome lengths with short spans dominant in the training sample. Unsurprisingly, we observe a trend of a performance increase as the frequency increases across the left hand plot with short outcomes, implying that the model struggles to recall infrequent outcomes despite their size but easily recalls the more frequent ones. 


\subsection{Random masking Vs custom masking}
\begin{figure}[h]
    \centering
    \includegraphics[width=\columnwidth]{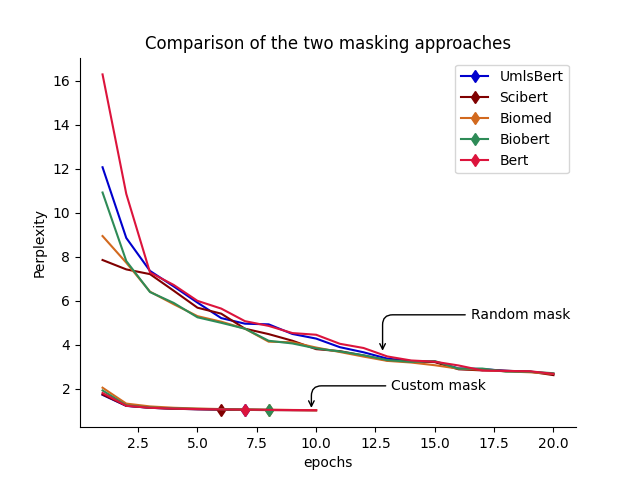}
    \caption{Achieving a target perplexity of 1.0 on the train dataset takes no fewer than 20 epochs with generic random masking of 15\% of the input prompt tokens \cite{devlin2018bert} compared to masking target factual information i.e. outcome spans themselves. Hitting target perplexity is shown using a diamond.}
    \label{fig:mask}.
     \vspace{-2em}
\end{figure}

\autoref{fig:mask} shows results of an ablation test in which we replace our custom masking approach with random masking. 
The key difference between the two is, while custom masking involves masking (or hiding) the outcomes in the prompts, random masking arbitrary masks 15\% of the prompts tokens.  
As shown in the figure, the number of epochs required to reach a perplexity of 1.0 on the train data for the two masking approaches is almost incomparable, with custom masking quickly achieving this in approximately 7 epochs and random masking failing to achieve this, even after 20 epochs. 
The earliest random masking achieves 1.0 perplexity is 80 epochs for SciBERT, however we only visualise 20 epochs because of space. Besides this, the insight suggests that, custom masking would significantly reduce GPU run-time or otherwise minimise overwhelming computational resources with massive datasets.

\begin{table*}[]
\centering
\resizebox{2\columnwidth}{!}{
\begin{tabular}{@{}llll@{}}
\toprule
\textbf{Query Variant}                                                                          & \textbf{Prompt}                                                                                                                                                         & \textbf{Correct}                                                                                   & \textbf{Generated outcomes}                                              \\ \midrule
\begin{tabular}[c]{@{}l@{}}\textbf{Cloze}\\ \{ctxt\} {[}M{]} \{ctxt\}\end{tabular}                        & \begin{tabular}[c]{@{}l@{}}Self-reported life-time medical diagnosis of \textbf{{[}M{]}} or use\\ of antidepressants was considered as outcome.\end{tabular}                     & - Depression                                                                                       & - Depression                                                             \\ \midrule
\begin{tabular}[c]{@{}l@{}}\textbf{Postfix}\\ {[}M{]} \{ctxt\}\end{tabular}                                  & \begin{tabular}[c]{@{}l@{}}\textbf{{[}M{]}} was assessed by questionnaires EORTC QLQ-C30, \\ and EORTC QLQ-BR23  at baseline, and at three, six,\\ and nine months.\end{tabular} & - Quality of life                                                                                  & - Life                                                                   \\
\midrule
\begin{tabular}[c]{@{}l@{}}\textbf{Prefix}\\ \{ctxt\} {[}M{]}\end{tabular}                                   & \begin{tabular}[c]{@{}l@{}}Two CMZ patients and one morphine patient showed\\ complete \textbf{{[}M{]}}.\end{tabular}                                                            & - pain                                                                                             & - unwanted pain                                                            \\
\midrule
\begin{tabular}[c]{@{}l@{}}\textbf{Mixed}\\ \{ctxt\} {[}M{]} \{ctxt\} \\ {[}M{]} \{ctxt\}\end{tabular} & \begin{tabular}[c]{@{}l@{}}Further additional benefits are better \textbf{{[}M{]}} and shorter \\ \textbf{{[}M{]}} compared with  standard GVHD prophlaxis\\  without ATLG.\end{tabular}  & \begin{tabular}[c]{@{}l@{}}- quality of life (QOL)\\ - immunosuppressive\\  treatment\end{tabular} & \begin{tabular}[c]{@{}l@{}}- immunosuppressive\\  treatment\end{tabular} \\ \cmidrule{2-4}
                                                                                                       & \begin{tabular}[c]{@{}l@{}}The incidence of postoperative \textbf{{[}M{]}}, \textbf{{[}M{]}}, \textbf{{[}M{]}} and \textbf{{[}M{]}}\\ was similar between the groups\end{tabular}                           & \begin{tabular}[c]{@{}l@{}}- nausea, - vomiting, \\ - drowsiness, -headache\end{tabular}           & - anxiety, - depression                                                     \\ \bottomrule
\end{tabular}
}
\caption{Example prompts from our test set and their predicted or generated outcomes for the outcome generation task. The Query variant column indicates the type of prompt as well as the prompt structure where \{ctxt\} implies context which might appear before, after or either ends of a masked sequence span.}
\label{tab:example_prompts}
 \vspace{-1em}
\end{table*}

\subsection{Error Analysis}
We analyse the outcomes generated by the best model (SciBERT+Contextual PBC+EBM-COMET) during the few shot evaluation and notice that whilst the model generates correct outcomes for some prompts, it makes various kinds of mistakes. 
\autoref{tab:example_prompts} includes a fair sample of the most commonly discovered mistakes. \textbf{Incomplete outcomes}, such in the Postfix where instead of ``Quality of life'', the model generates ``Life''. \textbf{Outcomes with irrelevant information}, such as Prefix case where the models generates more than what's expected, ``unwanted pain'' instead of ``pain''. Finally, \textbf{wrong outcomes}, where the model generates completely unexpected outcomes such as the case in the Mixed prompts.

\section{Related work}
Interrogating PLMs with fill-in-the-blank prompts to determine their knowledge and awareness of factual information is a trending paradigm in NLP. Despite the emergence of subtle techniques such as automating prompt structuring \cite{shin2020autoprompt, gao2020making}, selectively updating parameters of LMs and prompts (also known as continuous prompting) \cite{li2021prefix, qin2021learning}, or even not tuning at all \cite{brown2020language}, several works including these still heavily rely on handcrafted prompts to use in probing LMs. Our efforts are motivated by the fact that we need not worry about the nature of the prompt, but rather can leverage on information local to the prompt such as word positions to probe the LMs. 
We attempt to enhance a word's contextualised representation with position based representations to capture the word's position relative to the mask to be filled. Previously some works have used similar position-aware attention over LSTMs for relation extraction, sequence labelling and slot filling tasks in different datasets \cite{wei2021position, zhang2017position}. 
To the best of our knowledge, we are the first to use an extra position-attention layer above transformer models such as BERT to solve the fill-in-the-blank prompting task. 

\section{Conclusion}
This paper assesses the possibility of ignoring the constraint of aligning prompts to specific linguistic patterns in prompting tasks that aim to store knowledge in LMs that could later be retrieved or transferred for fact generation tasks. In experiments using clinical domain datasets (supporting EBM tasks), we show that the position-based attention implemented over contextualised LMs can improve the ability of PLMs to recall facts such as outcomes (biomedical entities) encountered during training. We further observe our proposed model is able to generalise across unseen prompts, performing considerably well for Cloze and Mixed (extremely rare in PBL tasks) prompts. With the obtained experimental results, despite not aligning our prompts to commonly followed linguistic patterns, we can positively answer the question posed in \textsection\ref{sec:Task} by claiming that PLMs are knowledgeable of stored facts.
\bibliographystyle{acl_natbib}
\bibliography{acl}

\section*{Appendices}
\appendix

\title{Position-based prompting for health outcome generation \\ -- Supplementary Materials --}

\date{}

\maketitle
\appendix
\section{Hyperparameters and Run time}
Using BioBERT in the Position based conditioning framework, we perform a grid search through multiple combinations of hyperparameters included in Table \autoref{tab:parameter_pbc} below. The model is tuned on 20\% of EBM-COMET dataset (as a dev set), we obtain the best Partial Match (PM) and Exact Match (EM) accuracies. Table \autoref{tab:parameter_pbc} shows the range of values (including the lower and upper bound) for which the model is tuned to obtain optimal configurations. Using a shared TITAN RTX 24GB GPU, the baseline model runs for approximately 40 minutes per epoch. 

\begin{table}[!htb]
\centering
\resizebox{\columnwidth}{!}{
\begin{tabular}{@{}lcc@{}}
\toprule
\textbf{Parameter}                                                             & \textbf{Tuned-range}                                                                & \textbf{Optimal}                                   \\ \midrule
Train Batch size                                                                     & {[}8,16,32{]}                                                                      & 16,32                                                 \\
Eval Batch size                                                                     & {[}8,16,32{]}                                                                      & 8                                                \\
Query type embedding size                                                                       & {[}50,100,150{]}                                                           & 50                                              \\
Position embedding size                                                                       & {[}100,200,300{]}                                                           & 300                                              \\
Attention layer size                                                                       & {[}100,200,300{]}                                                           & 200                                             \\
Optimizer                                                                      & {[}Adam, SGD{]}                                                                     & Adam                                               \\

Learning rate                                                                  & \begin{tabular}[c]{@{}c@{}}{[}5e-5, 1e-4, 5e-3, 1e-3{]}\end{tabular} & 5e-5                                               \\ \bottomrule
\end{tabular}
}
\caption{Parameter settings for the Position-based conditioning model}
\label{tab:parameter_pbc}
\end{table}

\section{Datasets}
\subsection{EBM-NLP}
EBM-NLP corpus~\cite{Nye2018ALiterature} is a crowd sourced dataset in which ca.5,000 clinical trial abstracts were annotated with elements in the health literature searching PICO framework~\cite{Huang2006EvaluationQuestions}. PICO stands for Participants, Interventions, Comparators and Outcomes. The dataset has supported clinicalNLP research tasks~\cite{beltagy2019scibert, brockmeier2019improving}. The corpus has two versions, (1) the ``\textbf{starting spans}'' in which text spans are annotated with the literal ``PIO'' labels (I and C merged into I) and (2) the ``\textbf{hierarchical labels}'' in which the annotated outcome ``PIO'' spans were annotated with more specific labels aligned to the concepts codified by the Medical Subject Headings (MeSH)~\footnote{\url{https://www.nlm.nih.gov/mesh}}, for instance the Outcomes (O) spans are annotated with more granular (specific) labels which  
include Physical, Pain, Mental, Mortality and Adverse
effects. For the clinical recognition task we attempt, we use the hierarchical version of the dataset. The dataset has however been discovered to have flawed outcome annotations~\cite{abaho2019correcting} such as (1) statistical metrics and measurement tools annotated as part of clinical outcomes e.g.``\textit{mean arterial blood pressure}'' instead of ``\textit{arterial blood-pressure}'',``\textit{Quality of life Questionnaire}'' instead of ``\textit{Quality of life}'' and (2) Multiple outcomes annotated as a single outcome ``\textit{Systolic and Diastolic blood-
pressure}'' instead of ``\textit{Systolic blood-pressure}'' and ``\textit{Diastolic blood-pressure}''.

\subsection{EBM-COMET}
A biomedical corpus containing 300 PubMed ``Randomised controlled Trial'' abstracts manually annotated with outcome
classifications drawn from the taxonomy proposed by~\cite{Dodd2018ADiscovery}. The abstracts were annotated by two experts with extensive experience in
annotating outcomes in systematic reviews of clinical
trials~\cite{abaho2021assessment}. \newcite{Dodd2018ADiscovery}'s taxonomy hierarchically categorised 38 outcome domains into 5 outcome core areas and applied this classification system to 299 published core outcome sets (COS) in the Core Outcomes Measures in Effectiveness (COMET) database.

\section{Layer probing}
Initially, the hidden state we used (Equation \eqref{eqn:plm}) extracted from the last layer for each of the Biomedical PLMs for all experiments. 
We however explore an option of extracting a weighted average of representation across all layers (Equation \eqref{eqn:w_plm}) as a hidden state and study the performance of the models once this hidden state is introduced in the Position based conditioning framework to obtain position-aware representations. 

\begin{align}
\label{eqn:w_plm}
\centering
\vec{h_i^{l}} = \mathrm{PLM_\theta}(x_i)
\end{align}
\begin{align}
\label{eqn:w_plm}
\centering
\vec{h_i} = \mathrm{MeanPool}(\vec{h_i^1,..,h_i^l,..,\vec{h_i^{l_N}}})
\end{align}
where $h_i^l$ is a hidden state extracted from the $l^{th}$ layer for word $x$ at position $i$.

We only repeat training experiments using the Contextual PBC setup (\autoref{sec:prediction}) however this time round using a mean pooled embedding across all layers as the hidden state. We notice that, aggregating a tokens representation by mean pooling across all layers of the transformer-based models does improve the performance in the outcome recalling experiments for both datasets. 

\begin{table}[!htb]
\centering
\resizebox{\columnwidth}{!}{
\begin{tabular}{@{}lcccc@{}}
\toprule
Dataset         & \multicolumn{4}{c}{EBM-COMET}                                                                                                                                                        \\ \midrule
Method          & \multicolumn{2}{c}{\begin{tabular}[c]{@{}c@{}}Contextual PBC\\ (last layer)\end{tabular}} & \multicolumn{2}{c}{\begin{tabular}[c]{@{}c@{}}Contextual PBC\\ (Mean pool)\end{tabular}} \\ \midrule 
Metric          & EM                                          & PM                                          & EM                                          & PM                                         \\ \midrule \midrule
BERT            & 43.32                                       & 55.94                                       & 45.80                                       & 57.19                                      \\
BioBERT         & 53.34                                       & 59.65                                       & 53.58                                       & 61.22                                      \\
SciBERT         & 63.00                                       & 70.95                                       & 63.15                                       & 72.67                                      \\
Biomed\_Roberta & 44.32                                       & 62.86                                       & 45.00                                       & 63.17                                      \\
UmlsBERT        & 31.88                                       & 36.46                                       & 33.10                                       & 39.21                                      \\ \midrule
Mean score      & 47.37                                       & 57.17                                       & 48.13                                       & 58.70                                      \\ \bottomrule
\end{tabular}
}
\caption{Table reports EM and PM accuracies of the various biomedical Pre-trained Language Models for the outcome recalling experiments using the EBM-COMET and Contextual PBC. Mean score in a particular column is the average across all results in that column.}
\label{tab:mean_pooling_layers}
\end{table}

\begin{table}[!htb]
\centering
\resizebox{\columnwidth}{!}{
\begin{tabular}{@{}lcccc@{}}
\toprule
Dataset         & \multicolumn{4}{c}{EBM-NLP}                                                                                                                                                          \\ \midrule
Method          & \multicolumn{2}{c}{\begin{tabular}[c]{@{}c@{}}Contextual PBC\\ (last layer)\end{tabular}} & \multicolumn{2}{c}{\begin{tabular}[c]{@{}c@{}}Contextual PBC\\ (Mean pool)\end{tabular}} \\ \midrule
Metric          & EM                                          & PM                                          & EM                                          & PM                                         \\ \midrule \midrule
BERT            & 47.31                                       & 51.06                                       & 47.45                                       & 53.41                                      \\
BioBERT         & 52.15                                       & 54.50                                       & 54.80                                       & 55.15                                      \\
SciBERT         & 59,44                                       & 63.91                                       & 60.08                                       & 66.93                                      \\
Biomed\_Roberta & 49.17                                       & 55.00                                       & 49.19                                       & 56.33                                      \\
UmlsBERT        & 39.16                                       & 40.15                                       & 41.12                                       & 42.41                                      \\ \midrule
Mean score      & 49.45                                       & 52.92                                       & 50.53                                       & 54.85                                      \\ \bottomrule
\end{tabular}
}
\caption{Table reports EM and PM accuracies of the various biomedical Pre-trained Language Models for the outcome recalling experiments using the EBM-NLP and Contextual PBC. Mean score in a particular column is the average across all results in that column.}
\end{table}

\end{document}